\documentclass{article}

\usepackage{microtype}
\usepackage{graphicx}
\usepackage{subfigure}
\usepackage{booktabs} 

\usepackage{hyperref}

\usepackage[accepted]{icml2024}

\usepackage{amsmath}
\usepackage{amssymb}
\usepackage{mathtools}
\usepackage{amsthm}

\usepackage{wrapfig}

\usepackage[font=small,labelfont=bf]{caption}

\usepackage{xcolor}
\definecolor{dark-blue}{rgb}{0.15,0.15,0.4}

\hypersetup{
    colorlinks=true,
    linkcolor=blue,
    filecolor=magenta,      
    urlcolor=cyan,
    citecolor=dark-blue,
    pdftitle={The No Free Lunch Theorem, Kolmogorov Complexity, and the Role of Inductive Biases in Machine Learning},
    pdfpagemode=FullScreen,
}

\newcommand{\diag}{\mathop{\mathrm{diag}}}

\usepackage[nameinlink, capitalize,noabbrev]{cleveref}

\theoremstyle{plain}
\newtheorem{theorem}{Theorem}[section]

\theoremstyle{definition}

\theoremstyle{remark}

\usepackage[textsize=tiny]{todonotes}

\usepackage[most]{tcolorbox}
\definecolor{lightblue}{HTML}{18282e}
\definecolor{lighterblue}{HTML}{f2fafd}
\newtcolorbox{abox}{colback=lighterblue,colframe=lightblue}

\NewTotalTCBox{\myverb}{ O{lighterblue} v !O{} }
{ fontupper=\ttfamily,nobeforeafter,tcbox raise base,arc=0pt,outer arc=0pt,
top=0pt,bottom=0pt,left=0mm,right=0mm,
leftrule=0.3mm,rightrule=0.3mm,toprule=0.3mm,bottomrule=0.3mm,boxsep=0.8mm,
colback=#1!10!lighterblue,colframe=#1!0!black,#3}{#2}

\usepackage{enumitem}

\usepackage{xcolor}

\icmltitlerunning{The No Free Lunch Theorem, Kolmogorov Complexity, and the Role of Inductive Biases in Machine Learning}

\begin{document}

\twocolumn[
\icmltitle{The No Free Lunch Theorem, Kolmogorov Complexity, \\and the Role of Inductive Biases in Machine Learning}



\icmlsetsymbol{equal}{*}

\begin{icmlauthorlist}
\icmlauthor{Micah Goldblum}{equal,school1}
\icmlauthor{Marc Finzi}{equal,school1}
\icmlauthor{Keefer Rowan}{school1}
\icmlauthor{Andrew Gordon Wilson}{school1}
\end{icmlauthorlist}

\icmlaffiliation{school1}{New York University}

\icmlcorrespondingauthor{Micah Goldblum}{goldblum@nyu.edu}
\icmlcorrespondingauthor{Marc Finzi}{maf820@nyu.edu}
\icmlcorrespondingauthor{Andrew Gordon Wilson}{andrewgw@cims.nyu.edu}

\icmlkeywords{Machine Learning, ICML}

\vskip 0.3in
]



\printAffiliationsAndNotice{\icmlEqualContribution} 

\begin{abstract}
No free lunch theorems for supervised learning state that no learner can solve all problems or that all learners achieve exactly the same accuracy on average over a uniform distribution on learning problems.  Accordingly, these theorems are often referenced in support of the notion that individual problems require specially tailored inductive biases. While virtually all uniformly sampled datasets have high complexity, real-world problems disproportionately generate low-complexity data, and we argue that neural network models share this same preference, formalized using Kolmogorov complexity.  Notably, we show that architectures designed for a particular domain, such as computer vision, can compress datasets on a variety of seemingly unrelated domains. Our experiments show that pre-trained and even randomly initialized language models prefer to generate low-complexity sequences.  Whereas no free lunch theorems seemingly indicate that individual problems require specialized learners, we explain how tasks that often require human intervention such as picking an appropriately sized model when labeled data is scarce or plentiful can be automated into a single learning algorithm.  These observations justify the trend in deep learning of unifying seemingly disparate problems with an increasingly small set of machine learning models.
\end{abstract}

\section{Introduction}

The problem of justifying inductive reasoning has challenged epistemologists since at least the 1700s \citep{hume1748philosophical}.  How can we justify our belief that patterns we observed previously are likely to continue into the future without appealing to this same inductive reasoning in a circular fashion?  Nonetheless, we adopt inductive reasoning whenever we learn from past experience.

More recently, in the late 1990s, \emph{no free lunch theorems} emerged from the computer science community as rigorous arguments for the impossibility of induction in contexts seemingly relevant to real machine learning problems \citep{wolpert1996lack, wolpert1997no}. One such no free lunch theorem for supervised learning states that no single learner can achieve high accuracy on every problem \citep{shalev2014understanding}. Another states that every learner is equally good in expectation over a uniform distribution on learning problems \citep{wolpert1996lack}.  Such a world would be hostile to inductive reasoning. The assumption that labelings are drawn uniformly ensures that training data is uninformative about unseen samples.

In contrast to this dismal outlook on machine learning, naturally occurring data involve structure that could be shared even across seemingly disparate problems. If we can design learning algorithms with inductive biases that are aligned with this structure, then we may hope to perform inference on a wide range of problems.  In this work, we explore the alignment between structure in real-world data and machine learning models through the lens of \emph{Kolmogorov complexity}.

The Kolmogorov complexity of an output is defined as the length of the shortest program under a fixed language that produces it. In \cref{sec:data}, we explain the connection between Kolmogorov complexity and compressibility. Virtually all data drawn from a uniform distribution as assumed by the no free lunch theorem of \citet{wolpert1996lack} cannot be significantly compressed, yet relevant real-world datasets are highly compressible. In particular, neural networks themselves can be used to create compressions of data labelings, upper bounding their Kolmogorov complexity.

We then demonstrate in \cref{sec:models} that modern neural networks also prefer low Kolmogorov complexity, complementing the low complexity of actual data.  While models implemented on a computer cannot generate data with complexity exceeding the length of their associated program, we find they actually prefer data that is far simpler.  We formulate simple languages for generating numerical sequences, under which we can directly measure the Kolmogorov complexity of a sequence. We use these languages to inspect the simplicity bias of both pre-trained and randomly initialized language models.  GPT-3 \citep{brown2020language} reliably favors less complex sequences, and bigger and better GPT-3 variants even more so. Notably, randomly initialized GPT models share this simplicity bias. 

To further emphasize the universality of this simplicity bias, we reshape tabular data from diverse domains, including click prediction and airline delay prediction, into images and feed them through convolutional computer vision architectures, showing that these vision architectures prefer correct labelings to random ones, even on data which do not remotely resemble natural images and have no spatial structure.  We then compute cross-domain generalization bounds via Kolmogorov complexity.

A common intuition associated with no free lunch theorems dictates that since a single learner cannot solve all problems, practitioners must inspect data and manually select an appropriate learner for the specific problem at hand.  For example, a practitioner might select a more constrained model to avoid overfitting on small datasets, or convolutional architectures to accommodate natural image data.  To the contrary, we show in \cref{sec:selection} that the meta learner which selects the best learning algorithm from cross validation suffers little from overfitting even when the number of models investigated is large, and the cost of selection is quickly overcome by gains in validation accuracy.

\begin{figure}[h!]
    \centering
    \includegraphics[width=1\columnwidth]{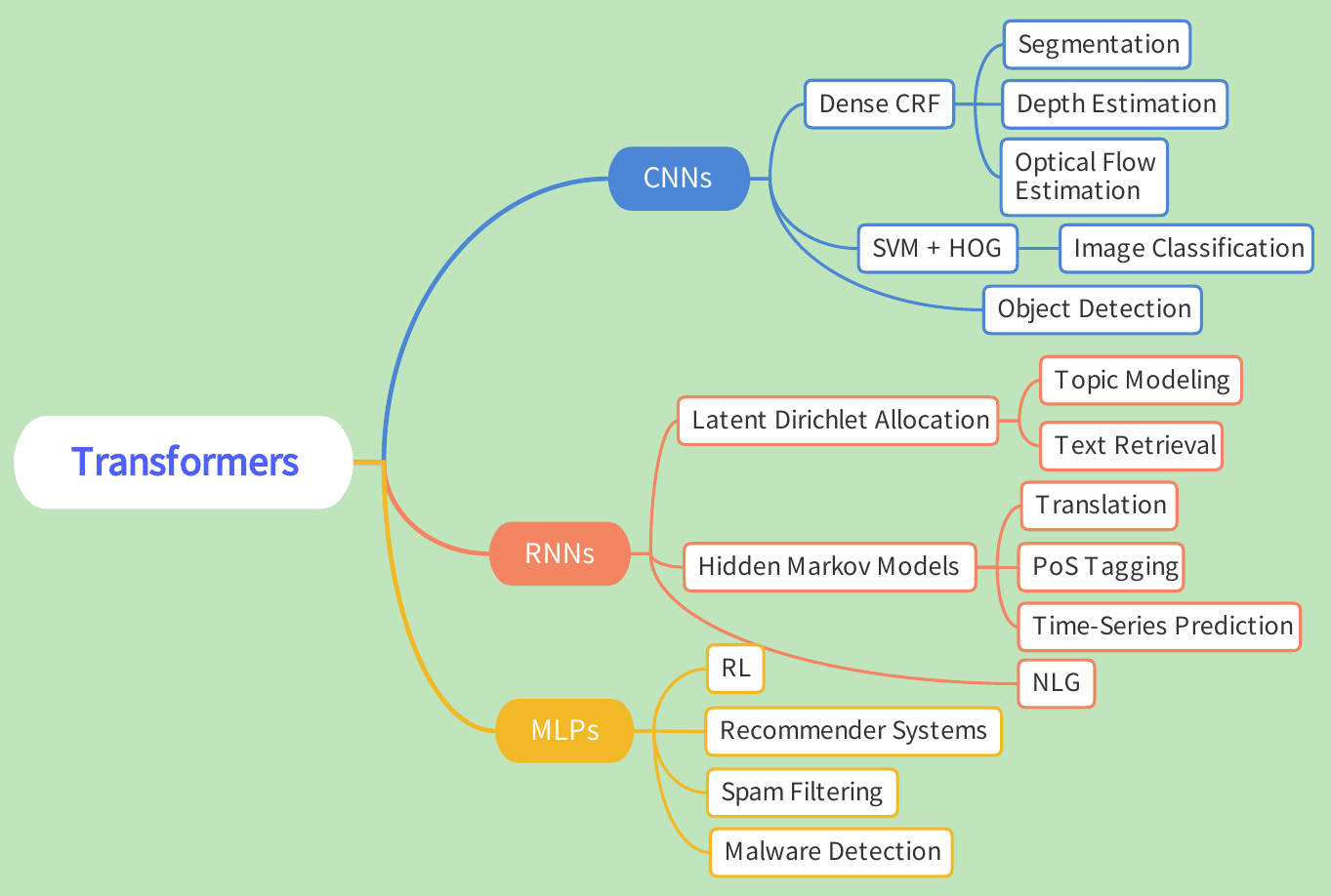}  
    \caption{Over time, tasks that were performed by domain-specialized ML systems are increasingly performed by unified neural network architectures. Real-world datasets often exhibit low Kolmogorov complexity. A model that combines a flexible hypothesis space with a simplicity bias towards low Kolmogorov complexity will provide good generalization on many different problems and modalities of data.}
    \label{fig:tree}
\end{figure}

Moreover, a single learner, which supports a variety of functions but prefers simple ones, can solve a wide range of problems.  We show that flexible models accompanied by a penalty encouraging simple solutions can solve problems at a variety of sample sizes.  
In fact, the historic evolution of machine learning supports the ability of a single learner to perform diverse tasks (see \cref{fig:tree}) as highly task-specific pre-neural algorithms, such as LDA \citep{blei2003latent} and HOG \citep{dalal2005histograms}, were replaced by neural architectures such as convolutional or recurrent models, and transformers can now handily perform all tasks listed in \cref{fig:tree}. We summarize some of our findings as follows:
\begin{itemize}[leftmargin=6mm]
\item We demonstrate the direct connection between compressibility and learnability that is implicit in no free lunch theorems by deriving a new no free lunch theorem using Kolmogorov complexity.
\item We show that the low Kolmogorov complexity of real datasets can be directly derived from the machine learning models used to fit them.
\item We compute the first cross-domain PAC-Bayes generalization bounds which show that neural networks such as convolutional architectures have low complexity biases that are relevant even on diverse tabular data far from what they were designed for.
\item We demonstrate GPT-3’s preference for sequences generated by short expression trees, finding even randomly initialized language models have a simplicity bias.
\end{itemize}
In short, while no free lunch theorems are regularly used to justify specially tailored inductive biases \citep[e.g.,][]{ho2002simple, whitley2005complexity, ciuffo2013no, watson1999algorithm}, we show that real-world data are not only highly structured, but share structure to a large extent. We further show how intervening to embrace a flexible hypothesis space together with a simplicity bias can lead to effective learners in small and large data regimes. Our findings also explain recently observed phenomena, ranging from the generality of transformers to the lack of overfitting on test sets of popular benchmarks noted in \citet{recht2019imagenet}. \textbf{This position paper argues that low-complexity structure shared by real-world datasets and machine learning models enables broad generalization across domains and sample sizes with a single model class.} We summarize key takeaways throughout the paper in \myverb{blue}.

\section{Background}
\label{sec:background}

We provide background on the no free lunch theorems, PAC-Bayes, and Kolmogorov complexity. We include an extended background discussion in \cref{app:extended_background}.

\textbf{No free lunch theorems.  }  No free lunch theorems (NFL) state that without making strong assumptions, a single algorithm cannot simultaneously solve all problems well. In supervised learning, the focus of this paper, \citet{wolpert1996lack} and  \citet{schaffer1994conservation} famously prove that
every learner---a function that takes in labeled data and outputs a labeling function for the associated domain---achieves the same average accuracy of $50\%$ on unseen examples over all binary classification problems. \citet{shalev2014understanding} instead do not assume a particular distribution over learning problems and prove that for every learner, there exists a task on which the learner achieves poor accuracy with high probability over training splits, whereas another learner achieves perfect accuracy.  Notably, the latter NFL computes accuracy over all data, not just ``off-training’’ samples.  The practical relevance of this theorem again hinges on the distribution over real-world learning problems and how well it aligns with the inductive bias of a learner.  In this paper, we argue that the real-world learning problems we care about share significant structure, and the inductive biases of neural networks are well-aligned with such problems. Note the distinction between the existence of learning problems where a learner generalizes poorly and out-of-distribution or adversarial test samples where a model fails to generalize.

\textbf{Kolmogorov complexity and compression.} Kolmogorov complexity quantifies the structure in a bitstring, measuring the extent to which it can be compressed. 
For a fixed programming language $L$, the Kolmogorov complexity of data $x$, $K(x)$, is the length of the shortest program in that language that outputs $x$ \citep{kolmogorov1963tables}. Analogous to conditional entropy, $K(y|x)$ is defined as the length of the shortest program which inputs $x$ and outputs $y$. Kolmogorov complexity provides a mathematical formalization of simplicity and Occam's razor, encompassing related concepts like Shannon information, compression, and minimum description length (MDL) \citep{li2008introduction}. While large Kolmogorov complexity is impossible to verify \citep{chaitin1974information}, all but exponentially few sequences of a given length have near maximal Kolmogorov complexity and are thus incompressible. Taken over the uniform distribution over bitstrings $x$, $P(K(x) \le n-k) \le 2^{1-k}$. However as we will discuss, these high complexity objects are rare in practice.

\textbf{Universal induction. } Inspired by Kolmogorov complexity, a line of work considers \emph{universal induction} methods, which prefer low complexity answers \citep{solomonoff1964formal, hutter2000theory, lattimore2013no, nakkiran2021turing, achille2021information}. Notably, Solomonoff induction \citep{solomonoff1964formal, rathmanner2011philosophical} makes predictions by applying Bayes rule to the universal prior which favors low complexity, and provides guarantees. Rather than formalizing theoretical learners that rely on Kolmogorov complexity, which is in general uncomputable, \citet{fernandez2014we} and \citet{gomez2016empirical} test popular machine learning algorithms on a diverse array of datasets to see if any existing algorithms are universal. 
Another line of work shows that a single transformer model can perform well on many problems \citep{mullertransformers, hollmann2022tabpfn}.

\textbf{PAC-Bayes generalization theory. } 
The PAC-Bayes framework is a convenient paradigm for proving generalization bounds on parametric models, while avoiding the pitfalls of uniform convergence. Rather than considering all elements of the hypothesis class on equal footing, we choose prior and posterior distributions over the parameters, and the generalization gap for elements of the posterior depends merely on the discrepancy between the two as measured by the KL divergence. This framework can explain many favorable properties of neural networks like flat minima \citep{hochreiter1997flat}, noise resilience \citep{arora2018stronger}, and compressibility \citep{zhou2018non}. It can also provide nonvacuous generalization bounds, with recent bounds drawing directly from Kolmogorov complexity and the universal prior \citep{kapoor2022}.

\textbf{On the relationship between our contributions and existing literature.} 
\textbf{(1)} In contrast to previous works which counter the no free lunch theorem by observing that a single model can achieve better-than-average empirical accuracy across diverse datasets \citep{fernandez2014we, gomez2016empirical}, we explain and formalize the structures which are universal across such data distributions using Kolmogorov complexity.  Relating this formalism to learning, we then show why low complexity is fundamental to such successes of machine learning models by proving a novel no free lunch theorem directly using Kolmogorov complexity. 
\textbf{(2)} The preference we demonstrate for low complexity emerges naturally in a variety of models, from transformer language models to convolutional neural networks, and requires no special interventions as proposed in \citet{schmidhuber1997discovering} or \citet{hinton1993keeping}.
\textbf{(3)} Existing generalization bound literature tunes priors on specific data distributions \citep{dziugaite2017computing, dziugaite2018data, perez2021tighter, dziugaite2021role} in line with the idea, often drawn from no free lunch theorems, that each domain requires a specially tailored model.  In contrast, we demonstrate that neural networks can compress a wide range of datasets in domains they were not even designed for, and that this compressibility can explain generalization via PAC-Bayes generalization bounds.
\textbf{(4)} Common wisdom dictates that neural network architectures must be carefully chosen for specific problems or sample sizes \citep{grinsztajn2022tree, brigato2021close, lee2021vision}, but we instead show through the formalism of complexity and experiments that specialized models can in principle be combined into a single learner which can perform well on a wide variety of problems and sample sizes. Moreover, we show that the cost of model selection is minimal, explaining recently observed phenomena such as a lack of overfitting to the test sets of popular benchmarks \citep{recht2019imagenet}.

\section{Unpacking the No Free Lunch Theorem with Kolmogorov Complexity}
\label{sec:data}

The often cited no free lunch theorem of \citet{wolpert1996lack} states that all learners perform the same when averaged over a uniform distribution on all possible datasets. However, since most possible datasets are incompressible, the assumption of uniform samples subtly selects high complexity incompressible data, where learning is fundamentally impossible. 
We elucidate the centrality of complexity in NFL theorems by deriving a new NFL theorem which uses the incompressibility of random data to show why on this data learning is impossible.  In \cref{app:dataset}, we provide a brief introduction to bounding the Kolmogorov complexity of a dataset by compressing it and including the file sizes of both compressed file and decompression code.  Through hypothesis testing, we rule out the possibility that real datasets are as high complexity as randomly drawn ones.

\subsection{NNs as Compressors of the Labeling Function}
\label{subsec:nncompressors}
Relevant to supervised learning, we show that not only are unlabeled datasets compressible---labeling functions are too. Further, we can  demonstrate their compressibility concretely using trained models as compressors. Given a labeled dataset $\mathcal{D} = (X,Y) = \{(x_i,y_i)\}_{i=1}^n$, any likelihood model $p(y|x)$---regardless of whether the top predictions are correct---can be used to generate a lossless compression scheme to encode the dataset labels $Y$ given the inputs $X$. Using a stream code such as arithmetic coding \citep{witten1987arithmetic}, in combination with the probability model $p(y|x)$, the labels can be encoded in $K(Y|X,p) \le -\sum_{i=1}^n\log_2 p(y_i|x_i)+2$ bits (see e.g. \citet{mackay2003information}). 
Models which maximize the log likelihood of the data also implicitly minimize the length of this encoding. 

\begin{figure*}[h!]
    \centering
    \begin{tabular}{ccc}

    \includegraphics[width=0.3\textwidth]{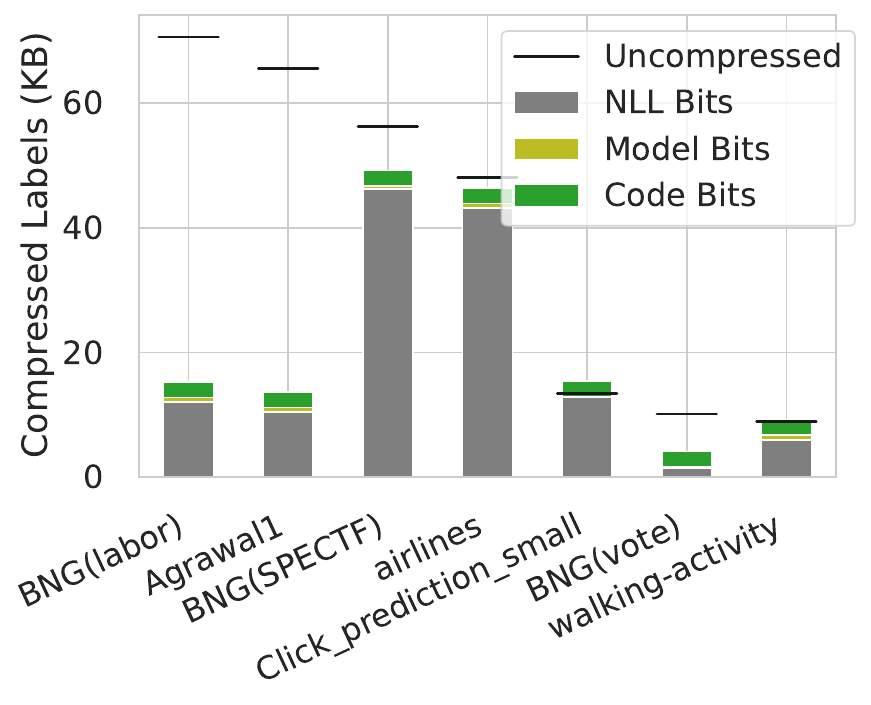} &
    \raisebox{7.1mm}{\includegraphics[width=0.3\textwidth]{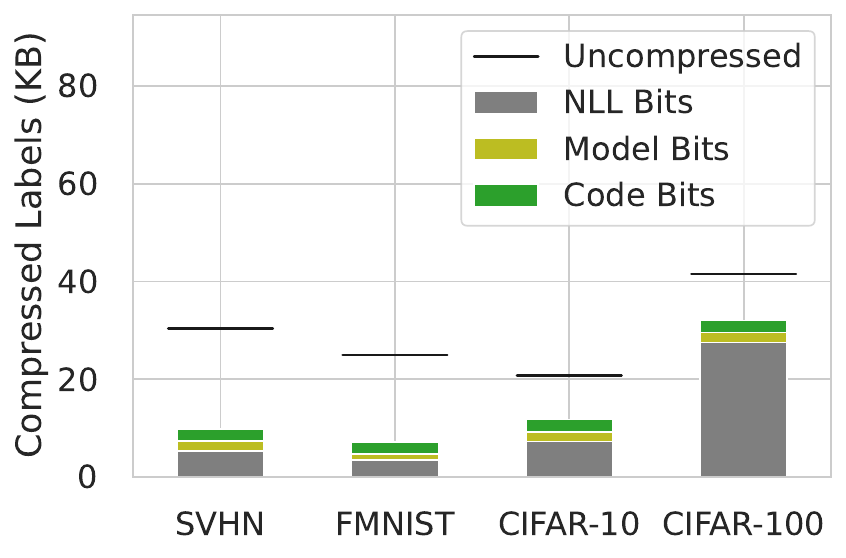}} &     \includegraphics[width=0.3\textwidth]{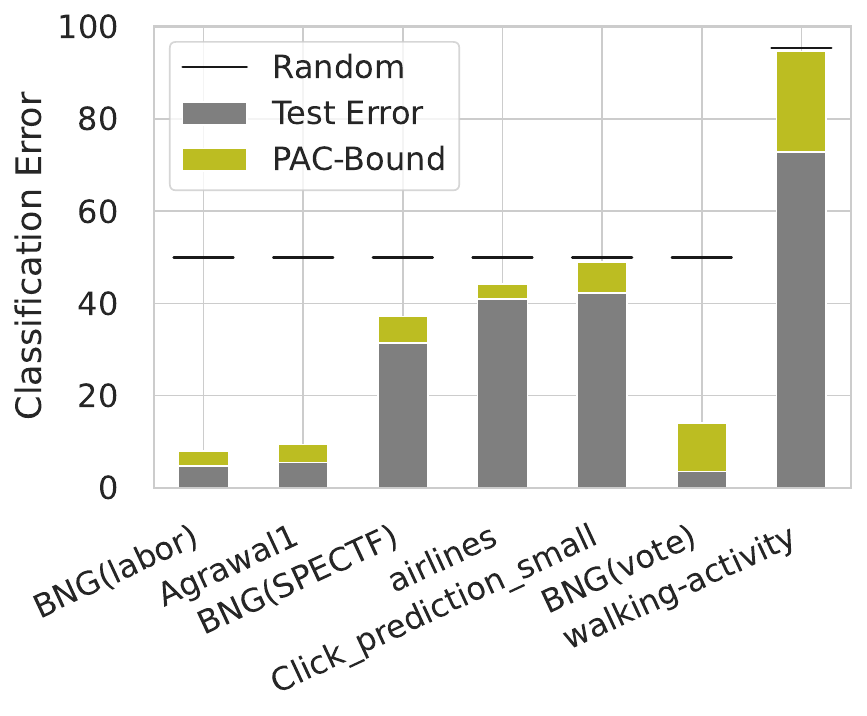}\\
    \end{tabular}
    \caption{(Left): Compressed sizes of tabular labels where compression is performed via a trained MLP model (as in \Cref{subsec:nncompressors}) vs. direct encoding of labels ($n\log_2 C$). (Middle): Compression of image classification datasets using CNNs. Note the breakdown of the total compressed size of the labels into model fit (NLL Bits), compressed parameters (Model Bits), and architecture and decompressor (Code Bits).  In both cases, models can greatly compress a diverse suite of datasets, highlighting a common structure shared by models and real-world data. (Right): Compression based generalization bounds \citep{kapoor2022} for CNNs on tabular data, fed in with each pixel representing a tabular feature. The bounds are able to explain the majority of the model performance as shown by the test error, indicating that even CNNs designed for computer vision have a generic inductive bias appropriate for a wide range of datasets containing no spatial structure at all.}
    \label{fig:pac-bounds}
\end{figure*}

As we derive in \autoref{app:nfl-proof}, $K(Y|X) \le K(Y|X,p)+K(p) + 2 \log_2 K(p) + c$, where $c$ is a small constant depending on the language. Writing the negative log likelihood in terms of the empirical cross entropy, combining our two inequalities, and  dividing by the size of the dataset $n$ yields
\begin{equation}
    \label{e.cond-kolmogorov-by-ce}
    \tfrac{1}{n}K(Y|X) \le \frac{{\mathrm{CE}}}{\ln 2} + n^{-1}(K(p) + 2\log_2 K(p) +c),
\end{equation}
where $\mathrm{CE}$ is the cross entropy of the classifier $p$ averaged over dataset $\mathcal{D}$. This inequality implies that, regardless of how large the model is, it provides a non-trivial compression of the dataset as the size $n$ of the dataset grows sufficiently large, as long as $\mathrm{CE}$ is better than random guess.  To demonstrate this fact, we employ the compression scheme from \citet{kapoor2022} in order to find a compressed representation of MLPs on several class balanced tabular classification datasets (available at \url{openml.org}). 
As shown in \autoref{fig:pac-bounds} (left), we are able to compress the labels on most of the datasets by well over the naive $n\log_2C$ encoding length where $C$ is the number of classes. We also apply the method with convolutional architectures to compress labels on CIFAR-10 and CIFAR-100 in \autoref{fig:pac-bounds} (middle), allowing us to reject the hypothesis that the labeling functions are drawn uniformly at random with extremely high confidence.

\subsection{A Kolmogorov-Style No Free Lunch Theorem}
A corollary of \autoref{e.cond-kolmogorov-by-ce} is that if the dataset is incompressible, then no model can do better than random chance in the large dataset limit, as we show in \cref{theorem}. Since compressible datasets in uniformly sampled data are exponentially unlikely, we can prove our own version of the no free lunch theorem. 
With very high probability, on any given uniformly sampled dataset, learning is impossible. 
\begin{theorem}
\label{theorem}
Let $(X,Y)$ be a dataset with $n$ data points and uniformly sampled random labels from $C$ classes. Then, with probability at least $1-\delta$, for every classifier $p(y|x)$,
\begin{equation}
    \mathrm{CE}(p) \geq \ln C - \frac{\ln 2}{n} \left(K(p) + 2 \log_2 K(p) + \log (1/\delta) + c\right),
\end{equation}
where CE$(p)$ is the empirical cross entropy of the classifier $p(y|x)$ on the data. Thus for any model of bounded size, if the size of the dataset is large enough, the model cannot represent any classifier with cross entropy appreciably smaller than that attained from random guess. Proof found in \autoref{app:nfl-proof}.
\end{theorem}
Like any of the no free lunch theorems, the necessary existence of unsolvable problems initially seems limiting. However, learning is in fact possible on compressible datasets (ones with less than maximal complexity).

\begin{abox}
Real datasets are highly unlike the high complexity samples from the uniform distribution, associated with no free lunch theorems, where learning is impossible.  The common structure shared by real datasets nullifies the limitations imposed by no free lunch theorems.
\end{abox}

\section{Low-Complexity Bias in Models}
\label{sec:models}

Previously, we saw that real-world data distributions across domains share a low Kolmogorov complexity bias.  If we can construct models which prefer low-complexity data, we can hope to perform inference with a single model across many domains.  While early machine learning systems incorporated domain-specific designs, such as handcrafted image features \citep{dalal2005histograms} or graphical models for language \citep{mnih2007three}, modern neural network architectures across domains are converging on transformers \citep{vaswani2017attention, dosovitskiy2020image, gulati2020conformer, somepalli2021saint}, some of which can simultaneously achieve impressive performance on a variety of data types with a single architecture \citep{jaegle2021perceiver}.

In this section, we argue that neural networks have a generic simplicity bias that extends beyond the datasets for which they are designed.  To this end, we: (1) feed tabular datasets from diverse domains such as click prediction and airline delay prediction into convolutional networks designed specifically for computer vision and find that they provably generalize well due to their simplicity bias, (2) formulate a language with respect to which we can measure the Kolmogorov complexity of numerical sequences and observe that GPT-3 generates low-complexity sequences with exponentially higher probability, (3) predict the next term in a sequence with randomly initialized language models.  Whereas the no free lunch theorem of \citet{wolpert1996lack} implies that such an inference procedure cannot outperform random guess on average, we find that randomly initialized neural networks prefer sequence completions which generate low-complexity completed sequences, demonstrating that they can make accurate guesses as long as the true sequence distribution also favors low complexity.
\subsection{Bounding Generalization by Complexity}
\label{subsec:pac}
Generalization bounds limit how the expected risk $R(h)$ for a model $h$ will differ from its train risk $\hat{R}(h)$. One simple such generalization bound is the finite hypothesis bound under a prior $P(h)$ \citep{langford2001bounds}: with probability $1-\delta$: $R(h) \le \hat{R}(h) + \sqrt{\frac{\log {1}/{P(h)} +\log {1}/{\delta} }{2 n}}$. 
Relating to Occam's razor and Solomonoff induction, consider the universal prior that assigns higher likelihood to compressible hypotheses: $P(h) = 2^{-K_p(h)}/Z$ where $K_p(h) \le K(h) + 2\log_2 K(h)$ is the \emph{prefix} Kolmogorov complexity and $Z \le 1$. Combining the two, we have with probability $1-\delta$,
\begin{equation}\label{eq:pac-bayes}
    R(h) \le \hat{R}(h) + \sqrt{\frac{K_p(h)\log 2 +\log {1}/{\delta} }{2 n}}.
\end{equation}
Despite the simplicity of the finite hypothesis bound, when combined with the universal prior, it provides 
nontrivial statements about 
generalization 
even for models which have many more parameters than data points \citep{kapoor2022}. Solutions found by many machine learning models on real datasets are highly compressible, and this reflects their bias for low Kolmogorov complexity functions.  Even under an arbitrarily large or even infinite hypothesis space, generalization is possible if we assign prior mass disproportionately to the highly structured data that typically occurs.

\subsection{Neural Networks Prefer Naturally Occurring Labelings Across Domains}
\label{subsec:crossdomain_pac}
The inductive biases of even specialized architectures like convolutional networks facilitate broad learning abilities. We now illustrate how a preference for low complexity alone is sufficient for a high degree of generalization, provably, since real-world data labelings tend to have low complexity.  To illustrate this fact, we take tabular classification datasets and encode the tabular features as an image by simply forming images where each pixel corresponds to a different feature, zero padding as necessary. We train a small convolutional network using this input data to predict the classification labels. Learning with the convolutional network requires overcoming the strong inductive bias tailored for the local and translation symmetric structure absent in this data. Even in spite of this extreme mismatch, the convolutional networks perform well. Using the compression and PAC-Bayes bound methodology from \citet{kapoor2022} (see \autoref{eq:pac-bayes}), we show the generalization bounds on these models along with test error in \autoref{fig:pac-bounds} (right). The strong generalization of convolutional networks on tabular datasets is almost entirely explainable through simplicity bias as the finite hypothesis bound nearly matches the test error.

\begin{abox}
Though CNNs were designed for vision, they generalize on unrelated tabular domains, a phenomenon almost entirely explained by their preference for low-complexity solutions.
\end{abox}

\subsection{GPT-3 Assigns Exponentially Higher Probability to Simpler Sequences}
\label{subsec:gpt3}

We now study the preference of GPT-3---a line of autoregressive LLMs---for simpler sequences. The ability of language models to solve reasoning problems has recently been studied by \citet{zelikman2022star}, who develop a prompting framework, and \citet{d2022deep}, who develop transformers for predicting symbolic expressions directly from the sequence. To perform our own study, we consider binary expression trees on basic arithmetic operations on integers, a simplified non-Turing-complete model of the larger space of programs. These expression trees can be executed to produce a numerical sequence. Complexity is defined with respect to unique encodings of these trees, with the value being the length of the shortest code for a tree which produces a given sequence as output. While distinct, the Kolmogorov complexity can be upper bounded by this complexity plus an added constant to encode the language. By using a small set of terms for the leaves and binary operators for the nodes, we can enumerate over all possible expression trees for small sizes with depth at most $L=7$ and compute all sequences with $0$ through $L$.

In our experiments, we use operations $+, \times,$ and  $//$ (integer division). For leaves, we use $2$ and $i$, where $i$ is the index within the sequence. For example, $(2+i)\times i$ could be implemented with a tree of size $2$ and would generate the sequence $a_i = 0, 3, 8, 15,...$  Using this setup, we generate sequences of varying complexity, according to a well-defined metric, and quantify the preference of GPT-3 models for simpler sequences. We provide details on tokenizing sequences and extracting their probabilities in \cref{app:gpt3}.

\begin{figure}[h!]
    \centering
    \includegraphics[width=0.49\textwidth]{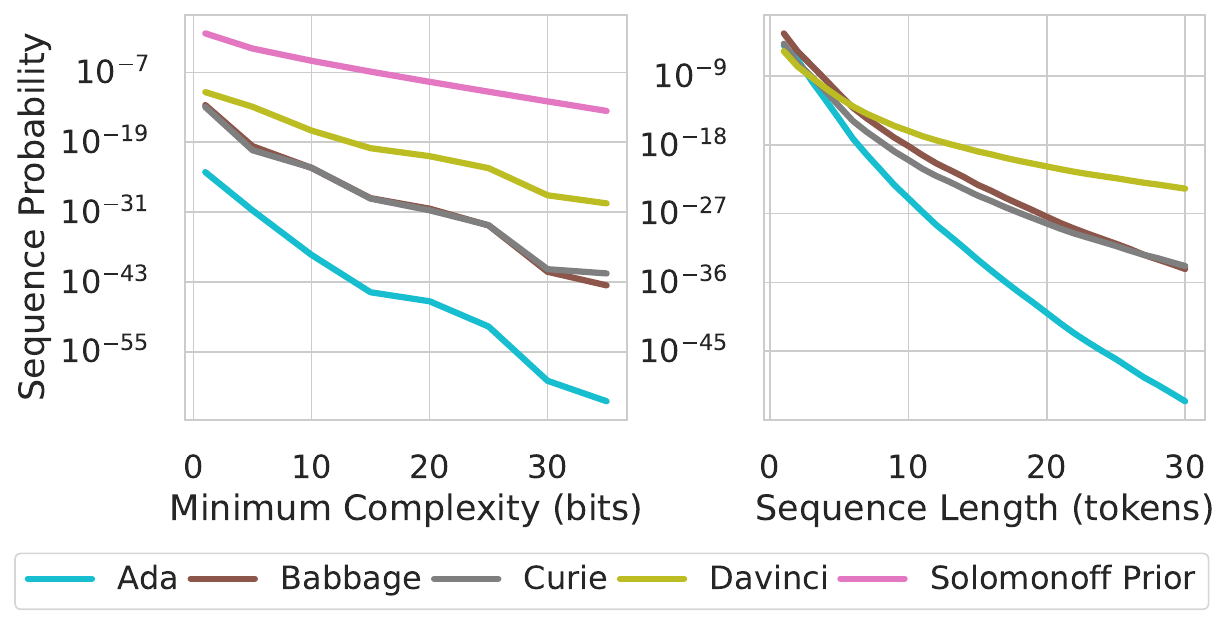}
    \caption{\textbf{GPT-3 prefers low-complexity sequences generated by expression trees. } \textbf{Left:} Average log-probability of sequences by complexity. \textbf{Right:} Average log-probability by sequence length, restricted to decimal digit tokens. GPT-3 variants ordered by increasing size.  Observe that GPT-3 variants assign exponentially lower probabilities to higher complexity sequences (left), as in the Solomonoff prior, and bigger more powerful models especially exhibit this behavior.  Moreover, the models become more confident as they see more tokens, and the more powerful GPT-3 variants such as \texttt{Davinci} learn faster (right).}
    \label{fig:gpt3}
\end{figure}

In \cref{fig:gpt3}, we measure the average log-probability GPT-3 models assign to sequences of a given complexity, where we fix the number of numerical tokens input into the model to be $30$, and we observe that the probabilities assigned by these language models decrease exponentially with sequence complexity, not far from the Solomonoff prior discussed in \cref{sec:background}.  In contrast, a uniform prior would be described by a flat line.  
We observe that big GPT-3 models which excel at language modeling, e.g. \texttt{Davinci} containing $175$ billion parameters, assign higher probability simple sequences than much smaller GPT-3 models such as \texttt{Ada}.

We can also examine the decay of such log-probabilities as we feed more digits of the sequence into the model. As the sequences get longer, we see in \cref{fig:gpt3} that the probabilities assigned to sequences decay sub-exponentially, indicating that these models, especially bigger variants, 
become increasingly confident about later sequence elements.

\subsection{Even Randomly Initialized Language Models Prefer Low Complexity}
\label{subsec:lm_init}

The previous section examined pre-trained language models, but these models were trained on massive corpora. Do they prefer low complexity at initialization before they have even seen any data at all?  While the initialization of neural network parameters is highly diffuse, these random parameters can induce a highly structured distribution over functions.  

Trained language models are known to repeat themselves \citep{holtzman2020curious, fu2021theoretical}.  One might think this behavior is learned from training data which contains repeated text, but we show randomly initialized GPT models repeat themselves too.  Interestingly, we can formalize the preference for repetition as an example of the broader preference for low Kolmogorov complexity.  To disentangle the impact of initialization from training, we adopt an even simpler setting which eschews the numerical tokens altogether. We consider binary sequences with arbitrary pairs of tokens constructed by simply repeating a given sequence until the output is of length $10$. Under this construction, the sequence $0, 0, 0, ...$ has complexity $1$, and $0, 1, 0, 1, ...$ has complexity $2$, yet randomly generated sequences are exponentially more likely to have high complexity.  We conduct our evaluations exhaustively on all such sequences of length $10$.

We now generate sequences of length $10$ with randomly initialized GPT-2 models \citep{radford2019language}, using each initialization to generate one sequence, and we measure the frequency with which each sequence is generated. We compare generation probabilities against complexity in \cref{app:init_lm} where we see again that low-complexity sequences are assigned exponentially higher probabilities.  Here, we compare (1) the uniform distribution over sequences, (2) randomly initialized GPT-2, as well as (3) pre-trained GPT-2 models.  We see that randomly initialized parameters induce a structured distribution over sequences, and pre-trained checkpoints exhibit an even stronger preference for low complexity as they are trained on structured text.  We can also use randomly initialized language models to perform next element prediction by estimating the probabilities they assign to the next element in a sequence given having correctly generated the previous terms.  While Wolpert's no free lunch theorem \citep{wolpert1996lack} ensures that the average completion accuracy over all possible length $10$ bitstrings is exactly $0.5$, we verify in \cref{app:init_lm} that randomly initialized networks can be used for sequence completion when the sequence has low complexity.

We can further generate long length-$100$ sequences with randomly initialized and pre-trained GPT-2 models and run a simple hypothesis test, demonstrating both randomly initialized and pre-trained models generate lower Kolmogorov complexity sequences on average than a uniform distribution.  We generate  $100{,}000$ samples from each of these three generative distributions and perform a one-tailed t-test on the null hypothesis that $\mu(K(S_{\text{GPT}}))\geq \mu(K(S_{\mathcal{U}}))$, where $S_{\text{GPT}}$ and $S_{\mathcal{U}}$ respectively denote random sequences generated by the language model or a uniform distribution.  Performing this test, we reject the null hypothesis in both randomly initialized and pre-trained models with an extremely low p-value, indicating that language models indeed prefer to generate simple sequences. 
Details are found in \cref{app:init_lm}.  We conclude that neural networks for language generation, both trained and randomly initialized, express a bias towards low Kolmogorov complexity which mirrors that of data as demonstrated in \cref{sec:data} and previously observed for classifiers in \citet{valle2018deep}.  Our findings also harmonize with \citet{huh2022low}, who show that even randomly initialized models express a preference for low effective rank embeddings. It is our contention that this simplicity bias leads to general-purpose learning.

\begin{abox}
Language models, both pre-trained and randomly initialized, prefer to generate low-complexity sequences.  As a result, we can use even such randomly initialized models to predict the next element in a sequence, as long as the sequence is low-complexity.
\end{abox}

\section{Model Selection with a Simplicity Bias}
\label{sec:selection}

In typical industrial workflows, practitioners examine their data and select an appropriate learner. 
We can then consider the human model selector and the model they select as a single meta-learner.  Whereas the no free lunch theorems seemingly preclude automated meta-learners which select performant models on any task, empirical works show that model selection can in fact be automated in practice  \citep{vilalta2002perspective}. 
 \citet{giraud2005toward} show that with minimal assumptions, the defeating conclusion of Wolpert's no free lunch theorem is escaped as long as datasets share structure so that the model selector generalizes to new datasets. In this section, we argue why in principle, model selection can be automated from the view of Kolmogorov complexity.

\subsection{Model Selection and Generalization Bounds}
\label{subsec:selection_bounds}
When developing a machine learning approach for an application, it is often helpful to leverage domain knowledge in constructing or choosing the right model for the task. One might start by choosing from families like MLPs, CNNs, GNNs, PointNets, or Transformers and then decide on the appropriate way of featurizing inputs, possibly incorporating knowledge of data symmetries via hard-coded equivariances or data augmentations. 
Even if we are extremely generous and suppose the practitioner is choosing from $100$ million models, we can consider the impractical algorithm of selecting one via cross validation. While one might expect that such a procedure would overfit, even finite hypothesis bounds show that it does not. Using cross validation on a validation set of size $20000$ for a classification problem, plugging in a uniform prior $P(h)=1/|\mathcal{H}| = 10^{-8}$ to the finite hypothesis bound in \Cref{subsec:pac}, we get that the gap between validation and test error will be less than $3.4\%$ with probability greater than $99\%$. Ultimately, we avoid overfitting because we only need a number of data points proportional to the log of the size of the hypothesis space. This reasoning can also be applied to theoretically resolve the empirical observation in \citet{recht2019imagenet} that we are not overfitting the test sets of popular benchmarks (more discussion in \cref{sec:discussion}).

For a more general class of models, consider the number of bits needed to specify model architectures like MLPs, CNNs, or GNNs, as well as symmetries and any other required information. In each case, the architecture can be expressed in few bits. A near state-of-the-art computer vision model can be expressed in only 280 characters \citep{trockman2022patches} in PyTorch. Similarly, important symmetries like translations, rotations, and other matrix groups can be expressed in few lines of code \citep{finzi2021practical} and can be used to encode equivariances or for augmentation. Therefore, even in selecting from all possible models that can be expressed in that short amount of code, we can expect to generalize with only tens of thousands of data points.

\begin{abox}
In principle, automating model selection directly via cross validation provably generalizes well across millions of models with only thousands of data points.
\end{abox}

\subsection{One Model for Big and Small Training Sets}
\label{subsec:one_model}

It is commonly believed that small training datasets demand compact architectures, whereas large training sets can accommodate flexible ones.  Accordingly, practitioners hand select appropriate models for their datasets.  We now show how we can intervene on the principle of combining flexibility with a simplicity bias, explored throughout the paper, to argue that a single learner can be effective for all data sizes. Our prior should prefer simple functions we believe are more likely yet support a wide variety of functions. We begin with a simple illustration on polynomial regression.

\begin{figure}[h!]
    \centering
    \includegraphics[width=0.42\textwidth]{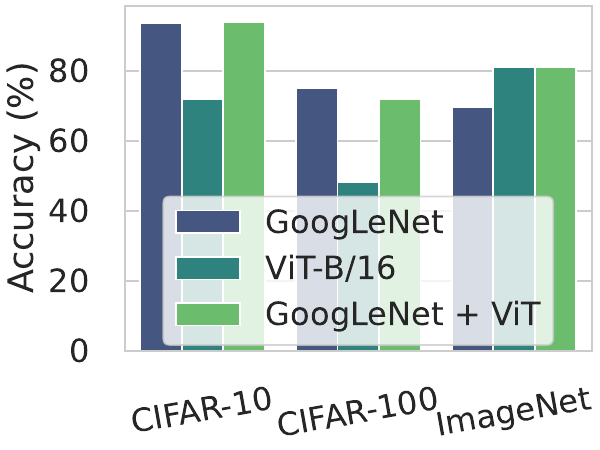}  
    \caption{A single learner, which is more expressive than a ViT but also prefers simple solutions representable by a GoogLeNet, can simultaneously solve small and large scale problems.}

    \label{fig:bigsmall}
    \vspace{-0.3cm}
\end{figure}

\textbf{Polynomial regression.  }  Common intuition dictates that high degree polynomials overfit small training sets.  In contrast, low degree polynomials cannot fit complicated functions so they should be avoided when training data is plentiful.  However, we find that a single high degree polynomial can be effective across sample sizes as long as we encode a preference for low-complexity solutions, which rely on low degree coefficients. 
 To this end, we adopt Tikhonov regularization with Tikhonov matrix $\diag(\{\alpha k^2\}_{k=0}^{d})$; in particular, we impose an $\ell_2$ penalty that increases quadratically with the degree of the corresponding monomial. In \cref{app:training_size}, we see that this model, which is flexible yet has a strong simplicity bias, performs at least on par with a low degree polynomial when training data is scarce, and with a high degree polynomial when training data is abundant.

\textbf{Neural networks.  }  We illustrate a similar concept with neural networks.  We consider a small network, GoogLeNet \citep{szegedy2015going}, which performs well on small datasets such as CIFAR-10 and CIFAR-100 \citep{krizhevsky_learning_2009}, but poorly on larger datasets like ImageNet \citep{deng2009imagenet}.  We also consider a large network, ViT-B/16 \citep{dosovitskiy2020image}, which performs significantly worse on CIFAR variants but better on ImageNet. We can similarly combine these two architectures, specifying a preference for GoogLeNet insofar as it fits the training data.  We train both models and then take a convex combination of their logits, $c *  \text{logits}_{\text{ViT}} + (1-c) * \text{logits}_{\text{G}}$, controlled by a parameter $c$ with $\ell_2$ regularization in favor of GoogLeNet (i.e., adding $\lambda c^2$ to the loss).  In \cref{fig:bigsmall}, we observe that while GoogLeNet and ViT each have strengths and weaknesses, combining them with a preference for simplicity achieves the best of both worlds. GoogLeNet and ViT can be combined into a single learner more flexible than GoogLeNet and with stronger simplicity bias than ViT, so that manual selection between them is not required across data size. 

In summary, flexible models with a low-complexity bias can be a one-stop-shop for machine learning since real-world data prefers low complexity. We do not need to compromise on flexibility in order to express a preference for low complexity solutions. Instead, follow Occam’s Razor and choose the simplest explanation for the training set.  We provide experimental details and additional experiments with Swin Transformer \citep{liu2021swin} in \cref{app:training_size}. Additional examination of complexity regularization across sample sizes, including linear models and neural networks, can be found in \citet{barron1991complexity} and \citet{nakkiran2020optimal}.

\begin{abox}
A single model can work well with both small and large training sets, so long as we embrace flexibility combined with a soft simplicity bias.
\end{abox}

\section{Discussion}
\label{sec:discussion}

In this section, we include a discussion of several fundamental themes that surface throughout the paper.

\textbf{Are the no free lunch theorems relevant to real-world model construction?} Not directly. They should not be used as an argument that we cannot significantly automate machine learning or science, as they so often are. The assumptions of these theorems --- such as a uniform distribution over all datasets --- are completely misaligned with the real world, where data are often highly structured (\cref{sec:data}). There is a valid discussion to be had about the role of inductive biases in model construction, but the no free lunch theorems should play no part. Our paper provides evidence, with PAC-Bayes bounds (\cref{subsec:crossdomain_pac}), and generative likelihoods (\cref{subsec:gpt3} and \cref{subsec:lm_init}), that the structure across many real-world datasets is shared to a surprising extent. These findings are aligned with the current movement towards similar transformer-based architectures for many tasks, spanning vision, NLP, and tabular data, and away from more specialized models for each task (\cref{fig:tree}). 

\textbf{How do we build models that are broadly applicable?} 
An emerging principle is that \emph{we should embrace a flexible hypothesis space, while providing soft encouragement towards salient structures} \citep{wilson2020bayesian}. In other words, we should avoid restriction biases, often represented by architectural constraints, such as parameter sharing and translation equivariance, in favour of softer inductive biases. 
We see this principle surfacing in many contexts. Transformers lack hard constraints but have recently been found to discover more equivariant solutions even than models with hard constraints \citep{gruver2022liederiv}. Outside deep learning, there is also work showing how soft inductive biases can be used to automate kernel selection on a variety of tasks which previously called for carefully hand-constructed kernels \citep{benton2019function, wilson2013gaussian, lloyd2014automatic}.
In this paper, we provide evidence that models can be made data efficient, while providing strong performance in larger data regimes, by embracing flexibility combined with soft inductive biases (\cref{sec:selection}). 

\textbf{Should the model (i.e. learner) we use depend on how much training data are available?} The conventional wisdom is yes. In principle, the answer is no. 
The \emph{conventional wisdom} is that a flexible model cannot be well-determined by a small training set and will overfit. This idea partly arises from early generalization theory, relying on Rademacher complexity \citep{mohri2009rademacher} or VC dimension \citep{vapnik1998adaptive}, and empirical overfitting of large models to small datasets. But there are many counterexamples; indeed, we often use flexible models with more parameters than datapoints, without explicit regularization, and achieve good generalization \citep{zhang2016understanding}. 
Such counterexamples predate deep learning. 
Indeed, examples of \emph{benign overfitting} in \citet{zhang2016understanding}, where a CNN fits randomly labeled images, can be reproduced by other model classes \citep{smith2017bayesian, wilson2020bayesian}. Moreover, there are generalization theories, such as PAC-Bayes, which do not penalize large hypothesis spaces, but instead focus on which solutions are a priori \emph{likely} \citep{mcallester1998some}.

\emph{In principle}, our beliefs about the generative process for data would not typically depend on how many points we happen to observe \citep{neal1996}. 
Typically, we would believe that there are many solutions that are a priori \emph{possible}, even if most of them are not a priori \emph{likely}. We should therefore embrace a large hypothesis space, regardless of how much data we have access to \citep{wilson2020bayesian}. 
We have shown several examples of how we can construct models that are competitive in both small and large data regimes, by embracing the principle of a flexible hypothesis space combined with soft inductive biases towards simplicity. 

\textbf{How crucial are the implicit biases of the optimization procedure in finding simple generalizable solutions?} The ability for deep models to generalize is often attributed to the implicit biases of stochastic optimizers like SGD or Adam \citep{amir2021sgd, wu2020noisy}. However, good generalization can be achieved by full-batch training, even without explicit regularization \citep{geiping2021stochastic, izmailov2021bayesian} or gradient-based optimization at all \citep{chiangloss}. Indeed, while stochastic optimizers can sometimes confer a small gain in performance, it is largely the design of the architecture that makes generalizable solutions more easily accessible (occupy a greater volume in the loss landscape), rather than the optimizer selecting for particularly generalizable low-loss solutions \citep{huang2020understanding}.

\textbf{Are we overfitting to benchmarks by comparing many models?}
Model development has been motivated by improving performance on benchmarks, leading to a concern that we may overfit to particular test sets. 
\citet{recht2019imagenet} find that the rankings of the best performing models is essentially preserved on new CIFAR and ImageNet test data, concluding that drops in performance are likely due to minor distribution shifts rather than overfitting. While this finding is often characterized as surprising, our calculation in  \Cref{subsec:selection_bounds} using the finite hypothesis bound from \Cref{subsec:pac} provides a theoretical explanation. Even comparing one hundred million models on a test set smaller than CIFAR or ImageNet, we would not expect a drop in test error by more than a few percent. In short, the danger of overfitting to standard benchmarks by checking the test accuracy of many different models is in fact provably very small. 

\textbf{Are transformers universal or are they limited?}
Despite the seeming convergence of neural network architectures towards transformers across modalities, transformers are limited.  Transformers, like virtually all standard architectures, are limited by their expenditure of bounded compute per input instance.  This property prevents transformers from implementing many low-complexity algorithms that would generalize well, and it limits the probability distributions they can represent via autoregressive generation \citep{lin2021limitations}.  \citet{zhang2024transformer} observe that transformers fail to solve simple problems with recursive structure.  \citet{liu2022transformers} show that transformers solve such problems inefficiently and thus learn shortcuts with poor generalization in practice.  Even with chain-of-thought prompting, transformers are not Turing complete due to their finite context length \citep{upadhyay2023turing}.  Recent works have instead built recurrent models that can ``think’’ longer to solve more complex problems \citep{schwarzschild2021can, bansal2022end}.  Future work may develop transformers similarly with adaptive compute and also with external memory.

\section*{Impact Statement}
This paper presents work with the goal of advancing the community's understanding of machine learning. This understanding may change the way we think about and practice machine learning, which may in turn have societal consequences. 
Therefore, our work may lead to consequences down the line, none which we feel must be specifically highlighted here.

\section*{Acknowledgements}
MG, MF and AGW are thankful for support from NSF CAREER IIS-2145492,
NSF CDS\&E-MSS 2134216, NSF HDR-2118310, BigHat Biosciences, Capital One, and an Amazon Research Award.

\bibliography{main}
\bibliographystyle{icml2024}

\newpage
\appendix
\onecolumn

\section{Extended Background}
\label{app:extended_background}

This section provides an extended version of the background presented in \cref{sec:background}.

\textbf{No free lunch theorems.  }   No free lunch theorems (NFL) state that without making strong assumptions, a single algorithm cannot simultaneously solve all problems well.  No free lunch theorems for search and optimization indicate that all optimizers and search algorithms satisfying certain conditions perform exactly the same on average over all such search or optimization problems \citep{wolpert1995no, wolpert1997no}.  In this work, we narrow our focus to NFL for supervised learning.  \citet{wolpert1996lack}, and similarly \citet{schaffer1994conservation}, proves an analogous theorem for supervised learning under which every learner---a function that takes in labeled data and outputs a labeling function for the associated domain---achieves exactly the same accuracy of $50\%$ on average over all binary classification problems where accuracy is only evaluated on unseen samples.

In order to prove no free lunch theorems, one needs to place very strong assumptions on the lack of structure in the labeling function, such as a uniform distribution, so that conditioning on the training labels does not modify the probability over labelings on unseen points \citep{rao1995every}.  To illustrate the severity of this condition, imagine being presented a sequence of one million $1$s and asked to predict whether the next element will be $1$ or $0$.  If labelings of sequence elements were distributed uniformly, then we should assign equal probability to both options, even though intuition and Bayesian probabilistic models tell us overwhelmingly to favor $1$. 

\citet{shalev2014understanding} instead do not assume a particular distribution over learning problems and prove that for every learner, there exists a task on which the learner achieves poor accuracy with high probability over training splits, whereas another learner achieves perfect accuracy.  Notably, the latter NFL computes accuracy over all data, not just ``off-training’’ samples.  While this statement of NFL does not explicitly require uniformly distributed data, the existence of catastrophic failure modes for our learners would not matter if our learners never encountered them in practice.  After all, we do not care if our learners cannot solve problems we do not want to solve.  Thus, the practical relevance of this theorem again hinges on the distribution over real-world learning problems and how well it aligns with the inductive bias of a learner.  Note the distinction between the existence of learning problems where a learner generalizes poorly and out-of-distribution or adversarial test samples where a model fails to generalize. In this paper, we argue that the real-world learning problems we care about share significant structure, and the inductive biases of neural networks are well-aligned with such problems.

\textbf{Kolmogorov complexity and compression.  } Kolmogorov complexity quantifies the structure in a bitstring, measuring the extent to which it can be compressed (an algorithmic definition of information content). For a fixed programming language $L$, the Kolmogorov complexity of data $x$, $K(x)$, is the length of the shortest program in that language that outputs $x$ \citep{kolmogorov1963tables}. Analogous to conditional entropy, $K(y|x)$ is defined as the length of the shortest program which inputs $x$ and outputs $y$. Kolmogorov complexity provides a mathematical formalization of simplicity and Occam's razor, which encompasses many related concepts like Shannon information, compression, and minimum description length (MDL) \citep{li2008introduction}.
 
While objects with large Kolmogorov complexity are impossible to verify \citep{chaitin1974information}, they are abundant over all possible bitstrings. All but exponentially few sequences of a given length have near maximal Kolmogorov complexity and are thus incompressible. Taken over the uniform distribution over bitstrings $x$, $P(K(x) \le n-k) \le 2^{1-k}$, where $n$ denotes the string length. However as we will discuss, these high complexity objects are extremely uncommon in practice.

\textbf{Universal induction. } Inspired by Kolmogorov complexity, a line of work introduces \emph{universal induction} methods, which prefer low complexity answers \citep{solomonoff1964formal, hutter2000theory, nakkiran2021turing}. Notably, Solomonoff induction \citep{solomonoff1964formal, rathmanner2011philosophical} makes predictions by applying Bayes rule to the universal prior which favors low complexity, and provides learning guarantees. The existence of universal learners calls into question the broader message of no free lunch theorems, showing that Occam's razor or a preference for low-complexity data labelings is sufficient for learning on low-complexity data \citep{lattimore2013no}.

Rather than formalizing theoretical learners that rely on Kolmogorov complexity, which is in general uncomputable, \citet{fernandez2014we} and \citet{gomez2016empirical} test popular machine learning algorithms on a diverse array of datasets to see if any existing algorithms are plausibly universal. \citep{wolpert1996lack} may not restrict machine learning in practice. Another line of work similarly shows that a single transformer model can perform well on a vast test bed of problems by mimicking Bayesian inference with in-context learning, and in fact achieves state-of-the-art on small tabular datasets \citep{mullertransformers, hollmann2022tabpfn}. 

\textbf{PAC-Bayes generalization theory. } 
The PAC-Bayes framework is a convenient paradigm for proving generalization bounds on parametric models, while avoiding the pitfalls of uniform convergence. Rather than considering all elements of the hypothesis class on equal footing, we choose prior and posterior distributions over the parameters, and the generalization gap for elements of the posterior depends merely on the discrepancy between the two as measured by the KL divergence. This framework can explain many favorable properties of neural networks like flat minima \citep{hochreiter1997flat}, noise resilience \citep{arora2018stronger}, and compressibility \citep{zhou2018non}. It can also provide nonvacuous generalization bounds, with recent bounds drawing directly from Kolmogorov complexity and the universal prior \citep{kapoor2022}.

Existing works in the generalization bound literature show that a model generalizes with high probability on a particular data distribution whenever it is compressible (i.e. has low complexity) with respect to the prior, but the prior is chosen specifically for the dataset at hand (e.g. CNNs for image classification) and furthermore the prior is often tuned directly on a fraction of the training set \citep{dziugaite2017computing, dziugaite2018data, perez2021tighter}.  In fact, there is a widely held belief in the generalization community that problem-specific priors, notably ones which are tuned on the training set, are necessary for strong generalization bounds \citep{dziugaite2021role}, and this belief manifests in data-dependent prior bounds across the literature.

In this paper, we argue against the necessity of problem-specific priors.  Our generalization bounds and compression experiments show that a single low-complexity biased prior can suffice on a wide variety of problems due to the low Kolmogorov complexity of data.  Whereas previous generalization theory literature is in line with the notion supported by no free lunch theorems that problems require specially tailored solutions, our work fights back against this widely held belief.

\textbf{Complexity in deep learning.}  Several works have related Kolmogorov complexity to neural networks.  One line of study proves that multi-layer perceptrons with Boolean input features are biased towards low-entropy functions, namely ones which classify disproportionately many or few points into the same class \citep{mingard2019neural} or are insensitive to flips in Boolean input features \citep{de2019random}. \citet{pearlmutter1990chaitin} argue that random initialization and noise in data increase the complexity of neural networks but that ensembling such models reduces complexity in expectation.  \citet{schmidhuber1997discovering} explicitly searches for simple neural networks with low Kolmogorov complexity and finds improvements in generalization on very small problems where such a search is computationally feasible. Other work measures the information contained in neural network activations \citep{achille2019information} or defines a complexity metric for datasets and uses this metric to predict transferability of pre-trained neural networks \citep{achille2021information}.  In another direction complimentary to Kolmogorov complexity, \citet{huh2022low} show that neural networks have a bias towards low effective rank Gram matrices computed from the deep features and that this simplicity bias increases with depth, explaining why bigger models can actually generalize better.

The compressibility of neural networks has also been studied for purposes other than PAC-Bayes generalization bounds.  For instance, \citet{blier2018description} explore the compressibility of datasets using neural networks, and show that variational methods yield poor compression whereas prequential coding can be used to obtain shorter code lengths.  \citet{hinton1993keeping} find that adding noise to neural network weights during training reduces the information contained in the weights, getting bits back and making the parameter vector compressible.  
Our work shows that the compressibility of datasets using neural networks is universal and does not require domain-specific models.

\textbf{On the relationship between our contributions and existing literature. } 
We summarize the relationship between our contributions and existing literature as follows:
\begin{itemize}[leftmargin=6mm]
\item In contrast to previous works which counter the no free lunch theorem by observing that a single model can achieve better-than-average empirical accuracy across diverse datasets \citep{fernandez2014we, gomez2016empirical}, we explain and formalize the structures which are universal across such data distributions using Kolmogorov complexity.  Relating this formalism to learning, we then show why low complexity is fundamental to such successes of machine learning models by proving a novel no free lunch theorem directly using Kolmogorov complexity. 
\item The preference we demonstrate for low complexity emerges naturally in a variety of models, from transformer language models to convolutional neural networks, and requires no special interventions as proposed in \citet{schmidhuber1997discovering} or \citet{hinton1993keeping}.
\item Existing generalization bound literature tunes priors on specific data distributions \citep{dziugaite2017computing, dziugaite2018data, perez2021tighter, dziugaite2021role} in line with the idea, often drawn from no free lunch theorems, that each domain requires a specially tailored model.  In contrast, we demonstrate that neural networks can compress a wide range of datasets in domains they were not even designed for, and that this compressibility can explain generalization via PAC-Bayes generalization bounds.
\item Common wisdom dictates that neural network architectures must be carefully chosen for specific problems or sample sizes \citep{grinsztajn2022tree, brigato2021close, lee2021vision}, but we instead show through the formalism of complexity and also through empirical experiments that specialized models can in principle be combined into a single learner which can perform well on a wide variety of problems and sample sizes. Moreover, we show that the cost of selection is minimal, explaining recently observed phenomena such as a lack of overfitting to the test sets of popular benchmarks \citep{recht2019imagenet}.
\end{itemize}

\section{An Exercise in Bounding Dataset Complexity}
\label{app:dataset}

We first consider the hypothesis that unlabeled machine learning datasets are drawn uniformly at random and use a bound on the Kolmogorov complexity as a test statistic. One can produce upper bounds on $K(x)$ by compressing $x$, but then it is necessary to include both the size of the compressed file and the size of the program required to decompress it.  Alternatively, if one can construct a short program which directly outputs $x$, this program also forms a compression of $x$.  Using \texttt{bzip2}, and including the size of the decompression program, we compress text dataset Amazon Review Full \citep{mcauley2013hidden} and audio dataset LibriSpeech \citep{panayotov2015librispeech} to 393.2 MB and 8.36 GB respectively, providing upper bounds on the Kolmogorov complexity with respect to the Python programming language. Computing the number of possible text and audio datasets of these sizes and supposing such datasets were in fact uniformly sampled at random, the probability of observing complexities this size or smaller is less than $10^{-1292913987}$ and $10^{-47632711550}$, astronomically low p-values, conclusively ruling out the possibility that they were sampled uniformly in this way. If we randomly shuffle the datasets, we instead obtain bounds of only 836.7 MB and 9.69 GB, considerably larger, showing that the compressibility results not just from an inefficient encoding, but from structure in the dataset.

Other works have also examined Kolmogorov complexity in data, for example EEG patterns \citep{petrosian1995kolmogorov} or animal behavior \citep{zenil2015approximations}, and confirm that such data is simple.  Our experiments above show that low Kolmogorov complexity is not specific to EEG patterns or animal behavior and is in fact a generic characteristic of common datasets we use in machine learning.  In this paper, we also use compressed neural networks to bound the complexity of labeling functions, rather than the model's inputs.

\section{A Kolmogorov No Free Lunch Theorem Proof}\label{app:nfl-proof}

\begin{theorem}
    With probability at least $1-\delta$ over datasets drawn from the uniform distribution, we have for every classifier $p$ over $C$ classes, the cross entropy is nearly as bad as random guess:
    \[\mathrm{CE}(p) \geq \ln C - \frac{\ln 2}{n} \left(K(p) + 2 \log_2 K(p) + \log_2 \delta + c\right)\].
\end{theorem}
\begin{proof}
Firstly, we relate the complexity of the classifier to the complexity of the labels $Y$ of the dataset:

\begin{align}
    K(Y|X) &\le K(Y,p|X)\\
    K(Y|X)       &\le K(Y|X,p) + K(p|X) + 2\log_2 K(p|X) +c \\
    K(Y|X)       & \le K(Y|X,p) + K(p) + 2\log_2 K(p) +c
\end{align}

For the second inequality, see e.g. \citep{fortnow2001kolmogorov}.

We can bound $K(Y|X,p)$ by coding the labels using $p$
\begin{equation}
    K(Y|X,p) \le -\sum_{i=1}^n\log_2 p(y_i|x_i)+2\le n\mathrm{CE}(p)+2
\end{equation}
where $\mathrm{CE}(p)$ is the empirical cross entropy (see e.g. arithmetic coding in \citet{mackay2003information}).

Rearranging, we have

    \[ \mathrm{CE}(p) \geq \frac{\ln 2}{n} \left(K(Y|X) - K(p) - 2 \log_2 K(p) -c\right).\]
    Note that by simply counting all possible programs taking input $X$, there are less than $2^{k+1}$ labelings $Y$ with $K(Y|X) \leq k$. Note that there are $C^n$ distinct labelings, from which we are drawing uniformly. So that
    \begin{align*}
    \mathbb{P}(K(Y|X) > n\log_2 C- m) &= 1 - \mathbb{P}(K(Y|X) \leq n\log_2 C -m) 
    \\&\geq 1 - \mathbb{P}(K(Y|X) \leq \lceil n\log_2 C\rceil -m)
    \\&\geq 1 - \frac{2^{\lceil n\log_2 C \rceil -m+1}}{C^n}
    \\&\geq 1 - 2^{2-m}.
    \end{align*}
    Alternatively, with probability at least $1-\delta$,
    \[K(Y|X) > n \log_2 C - \log_2 \frac{1}{\delta} -3.\]

Therefore:
\[\mathrm{CE}(p) \geq \ln C - \frac{\ln 2}{n} \left(K(p) + 2 \log_2 K(p) + \log_2(1/\delta) + c\right)\]
\end{proof}
\section{PAC-Bayes Compression Experimental Details}\label{app:pac}
For the OpenML tabular classification datasets, we preprocess them first by balancing the classes, subsampling all classes down to the number of examples of the least likely class. This way, when compressing the datasets (specifically, the labeling function), any result achieved is nontrivial in contrast with a very class imbalanced dataset.
We heavily follow the compression method of \citep{kapoor2022}, including the small $9$ convolutional architecture which they use to generate their bounds.
When cramming the tabular data into this convnet, we combine numerical features with one hot encoded categorical features and then pack these into the pixels of a $1$ channel image, using however large an image as necessary to fit each of the different features inside.

With respect to the sizes of the random subspaces that we train the compressed models in, we consider $250$,$500$,$1000$, and $2000$.
For tabular label compression, we employ a 2 hidden layer MLP with hidden dimension $k=192$, and we consider the same $250$,$500$,$1000$, and $2000$ values for subspace dimension. We train for $80$ epochs with $20$ epochs of quantization at a batch size of $512$ using Adam at lr$=3\times 10^{-4}$.
For image classification label compression, we use the $9$-layer convnet with subspace dimensions $2000$, $3000$, $5000$, and we train for $80$ epochs using SGD at learning rate $0.1$ and quantize for the remaining $20$ epochs, at a batch size of $50$.
For calculating the length of the code for model architecture and decompressor, we need only the implementation of the model, the arithmetic decoder, and the loading of the quantized values. Removing wasted bits, we minified the python file, leading to a size of approximately $2.5$KB.

\section{GPT-3 Experimental Details}
\label{app:gpt3}

To feed sequences into a model, we split up sequence elements into individual byte-pair encoding tokens corresponding to their decimal digits, and we place comma tokens between sequence elements as delimiters, also beginning every input with an \texttt{<|endoftext|>} token.  We choose to use the byte-pair encoding of decimal digits with a space inserted before the digit, e.g. ` 0' as this is known to enhance the ability of language models to perform arithmetic \citep{zelikman2022star}.  For example, the sequence $10, 11$ will be split up into [`\texttt{<|endoftext|>}', ` 1', ` 0', `,', ` 1', ` 1'], and each element of the list is tokenized individually. 
 Then, the log-probability of a sequence is given by the sum of the log-probabilities corresponding to the correct decimal digits in their respective slots of the model's output.  Note that various sequences will contain different numbers of decimal digits, and the sequence's log-probability will decrease with every token.  Therefore, in order for fair comparison, we limit all sequences to $30$ decimal digit tokens and truncate there.

\section{Sequence Generation and Completion with Randomly Initialized Language Models}
\label{app:init_lm}

For these experiments, we use Huggingface\footnote{\url{https://huggingface.co/}} GPT-2 architectures and pre-trained checkpoints.  In order to estimate the probabilities assigned by randomly initialized language models to each bitstring, we generate one million random sequences, ensuring that many instances of each bitstring are generated as there are only $2^{10}=1024$ bistrings of length $10$.

\begin{figure*}[h!]
    \centering
    \includegraphics[width=0.8\textwidth]{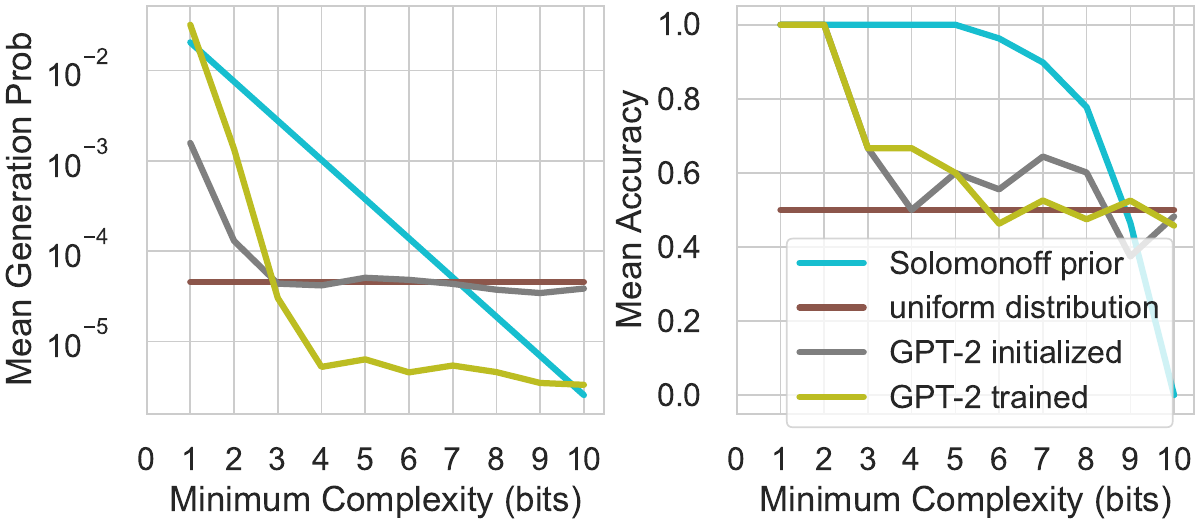}
    
    \caption{\textbf{Randomly initialized GPT-2 Base prefers low-complexity sequences generated by bitstring repetition. } \textbf{Left:} Average log-probability of sequences by complexity. \textbf{Right:} Average accuracy by complexity. 
    }
    \label{fig:gpt_init}
\end{figure*}

\begin{figure*}[h!]
    \centering
    \includegraphics[width=0.8\textwidth]{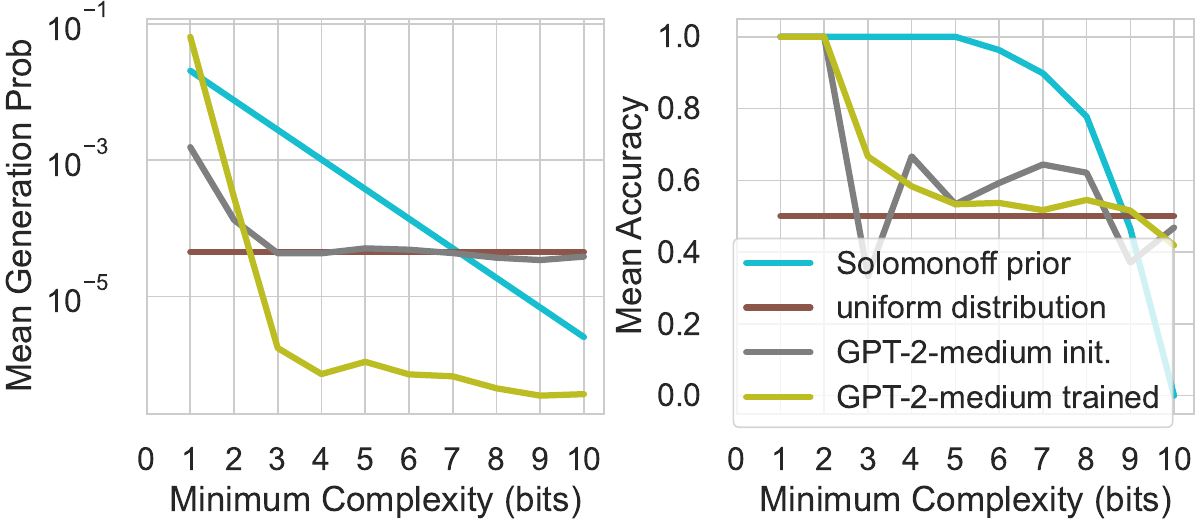}
    
    \caption{\textbf{Randomly initialized GPT-2 Medium prefers low-complexity sequences generated by bitstring repetition. } \textbf{Left:} Average log-probability of sequences by complexity. \textbf{Right:} Average accuracy.
    }
    \label{fig:gpt2_medium}
\end{figure*}

\begin{figure*}[h!]
    \centering
    \includegraphics[width=0.8\textwidth]{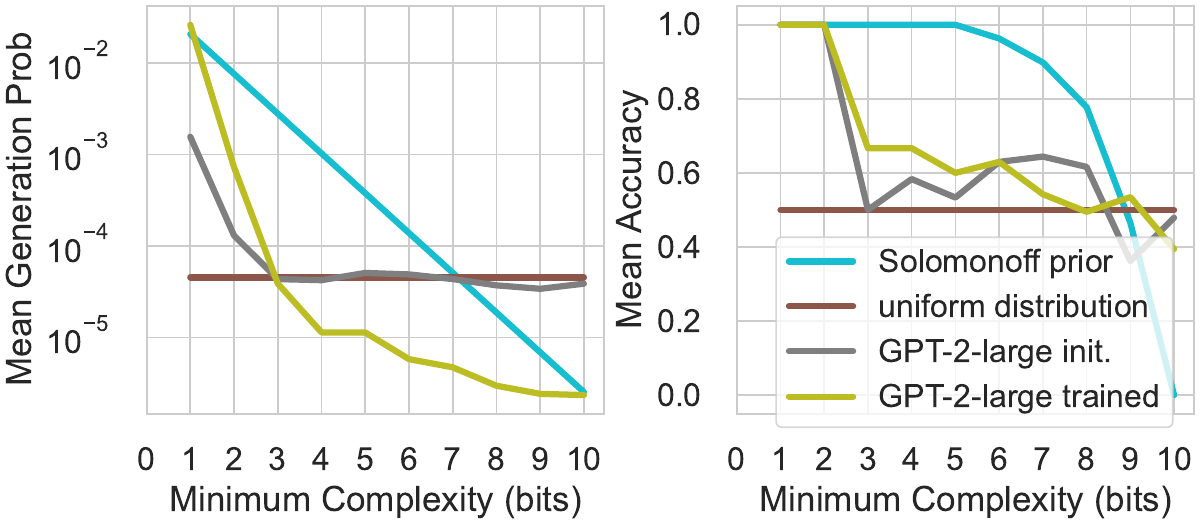}
    
    \caption{\textbf{Randomly initialized GPT-2 Large prefers low-complexity sequences generated by bitstring repetition. } \textbf{Left:} Average log-probability of sequences by complexity. \textbf{Right:} Average accuracy.
    }
    \label{fig:gpt2_large}
\end{figure*}

We include here plots with the various sizes of GPT-2 architectures in \cref{fig:gpt_init}, \cref{fig:gpt2_medium}, and \cref{fig:gpt2_large}.

We additionally include the hypothesis test referenced in \cref{subsec:lm_init}.  For this experiment, we generate $100{,}000$ length-$100$ sequences from randomly intialized GPT-2 variants, pre-trained GPT-2 variants, and a uniform distribution.  We then perform a one-sided t-test on the null hypothesis that $\mu(K(S_{\text{GPT}}))\geq \mu(K(S_{\mathcal{U}}))$, for both initialized and pre-trained models.  \cref{tab:ttest} contains the resulting sample means, t-statistics and p-values.  In all cases, we reject the null hypothesis with very low p-values, indicating that language models do prefer to generate low-complexity sequences.  Notably, pre-trained language models exhibit an increased simplicity bias, and bigger and better language models even more so.

\begin{table}[!h]
\caption{\textbf{Hypothesis test for language model simplicity bias.} t-tests are one-sided, and p-values are rounded to $4$ digits.  We also report the mean Kolomogorov complexity of sequences generated by each language model and a uniform distribution.}
\label{tab:ttest}
\centering 
    \begin{tabular}{llll}
    \toprule
        Model & $\overline{K(S_{\text{GPT}})}$ & t-statistic & p-value  \\
        \midrule
        Uniform Distribution & 98.36 & - & - \\

        GPT-2 Base Initialized & 98.00 & -39.95 & 0.0000 \\
        GPT-2 Medium Initialized & 97.99 & -40.91 & 0.0000 \\
        GPT-2 Large Initialized & 98.00 & -40.11 & 0.0000 \\

        GPT-2 Base Trained & 60.81 & -255.17 & 0.0000 \\
        GPT-2 Medium Trained & 48.41 & -325.16 & 0.0000 \\
        GPT-2 Large Trained & 46.34 & -342.80 & 0.0000 \\
    \bottomrule
    \end{tabular}
\end{table}

\section{Big and Small Training Sets}
\label{app:training_size}

\textbf{Polynomial regression.  } We choose three example target functions on which to perform regression: $\cos(\frac{3\pi}{2}x)$, $x^2$, and $-36x + 49x^5 - 14 x^7 + x^{10}$.  Training data is randomly drawn from a uniform distribution over the unit interval, and we add noise to training labels from $\mathcal{N}(0,0.1)$.  In each case, for each dataset size, we average the mean squared error over $100$ randomly sampled training sets.  For Tikhonov regularized polynomial regression on the cosine and degree 2 polynomial target functions, we use $\alpha = 0.01$, and we use $\alpha = 0.001$ for regression on the degree 10 polynomial target function.

\begin{figure*}[t!]
    \centering
    \begin{tabular}{ccc}
    \includegraphics[width=.95\textwidth]{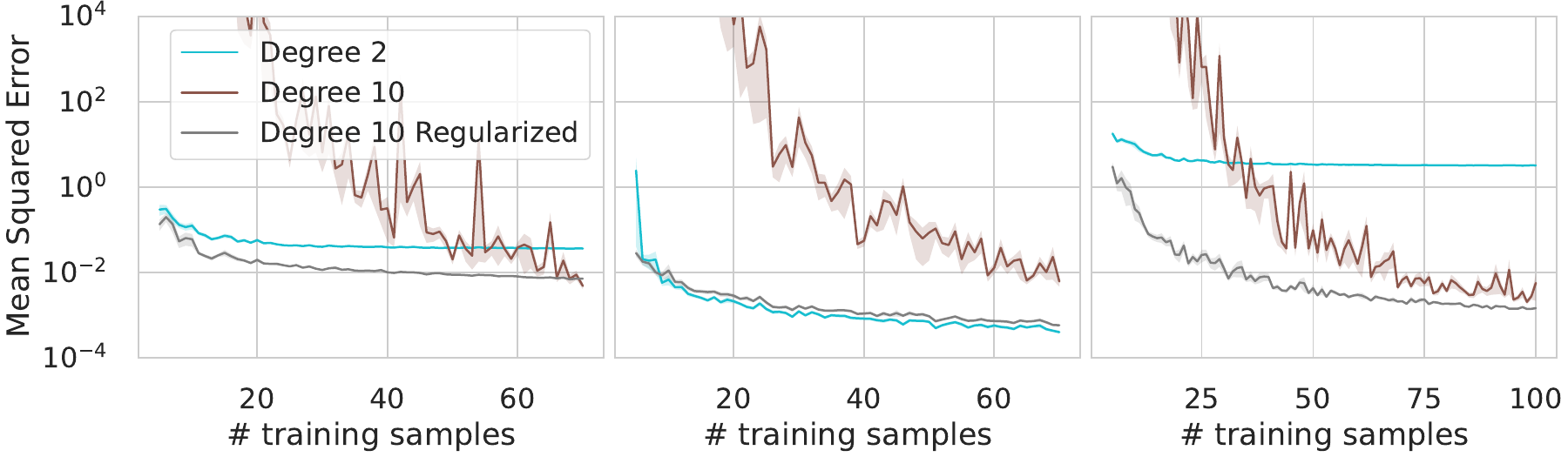}
    \end{tabular}
    \caption{\textbf{High-order polynomials with a complexity penalty can solve problems at a variety of sample sizes. } \textbf{Left:} Cosine target function. \textbf{Middle:} Degree $2$ polynomial target function. \textbf{Right:} Degree $10$ polynomial target function.}
    \label{fig:polynomial}
\end{figure*}

\textbf{Image classification with neural networks. }  For ImageNet trained models, we employ publicly available checkpoints from \texttt{torchvision}\footnote{\url{https://pytorch.org/vision/stable/index.html}}.  We train models on CIFAR-10 and CIFAR-100 for $200$ epochs with initial learning rate $0.1$ and cosine annealing along with horizontal flip and random crop augmentations.  We use SGD with momentum 0.9 and batches of size $128$.  All CIFAR images are rescaled to $224 \times 224$ so that we can use an identical model for ImageNet and CIFAR data.  In order to learn the parameter $c$ controlling the convex combination of models, we perform $10$ epochs of training, where the models' parameters are frozen, and we apply weight decay with coefficient $10^{-5}$.  We learn the parameter $c$ using SGD with momentum 0.9 and batch size $128$, initial learning rate 0.1, and cosine annealing.

\begin{table}[!h]
\caption{Combinations of large and small architectures form single models that achieve high test accuracy on all dataset sizes.  ``GoogLeNet + ViT'' denotes a model formed as a convex combination of the logits of the two constituent models with weight decay on the parameter $c$ controlling the convex combination which multiplies the logits of the larger model, ensuring that the small model is preferred as long as it fits the data.}
\centering
\begin{tabular}{llll}
\toprule
Model & CIFAR-10 & CIFAR-100 & ImageNet\\
\midrule
GoogLeNet & 93.840 \% & 75.160 \% & 69.778 \% \\
ViT-B/16 & 72.020 \% & 48.140 \% & 81.072 \% \\
Swin-B & 74.710 \% & 64.200 \% & 83.582 \% \\
GoogLeNet + ViT & 93.860 \% & 71.990 \% & 81.090 \% \\
GoogLeNet + Swin & 93.760 \% & 75.360 \% & 83.150 \% \\

\bottomrule
\end{tabular}
\label{tab:seq_init_genprob}
\end{table}

\section{Limitations}
In multiple experiments in this paper, we bound Kolmogorov complexity by compressing datasets or models.  These upper bounds on Kolmogorov complexity are likely very loose since our compressions probably are far from optimal.  We expect that high-performance models and the distribution over real-world datasets possess a much more severe simplicity bias than we can prove.  Moreover, a number of the experiments in this paper should be viewed as proof of concept.  For example, the compressibility of various datasets using neural networks provides evidence that real-world datasets are highly non-uniform and share a generic structure in common with neural networks. However, we can’t be sure that these observations will hold for all datasets and models as facts about the distribution of real-world datasets cannot be proven mathematically.

\section{Computational Resources}
In order to perform all experiments in this paper, we used a total of approximately $200$ GPU hours on NVIDIA RTX A4000 and NVIDIA Titan RTX cards.


\end{document}